\documentclass[letterpaper, 10 pt, conference]{ieeeconf}  

\IEEEoverridecommandlockouts                              

\usepackage{times} 
\usepackage{bbding}
\usepackage{graphicx}
\usepackage{float}
\usepackage{diagbox}
\usepackage{arydshln} 
\usepackage{bbm}
\usepackage{wrapfig}
\usepackage{mathrsfs}
\usepackage{listings} 
\usepackage{physics}

\usepackage{graphics} 
\usepackage{float}
\usepackage{multirow}
\usepackage{amsmath}
\usepackage{bbm}
\usepackage{amssymb}
\usepackage{mathtools}
\usepackage[utf8]{inputenc} 
\usepackage[T1]{fontenc}    
\usepackage{hyperref}       
\usepackage{cleveref}
\usepackage{url}            
\usepackage{booktabs}       
\usepackage{amsfonts}       
\usepackage{nicefrac}       
\usepackage{microtype}      
\usepackage{xcolor}         
\usepackage{algorithm}
\usepackage{algorithmicx}
\usepackage{algorithm} 
\usepackage{algpseudocode}
\usepackage{fancyvrb}
\usepackage{fvextra}
\usepackage{csquotes}
\usepackage{tcolorbox}
\usepackage{multicol}
\tcbuselibrary{skins}

\usepackage{threeparttable} 

\title{\LARGE \bf
NaviDiffusor: Cost-Guided Diffusion Model for Visual Navigation
}

\author{
Yiming Zeng*, Hao Ren*, Shuhang Wang, Junlong Huang, Hui Cheng
\thanks{This work was supported by the National Natural Science Foundation of China (U22A2095). Corresponding to chengh9@mail.sysu.edu.cn}
\thanks{* equal contribution. }
\thanks{Yiming Zeng, Hao Ren, Shuhang Wang, Hui Cheng are with the School of Computer Science and Engineering, Sun Yat-sen University.}
\thanks{Junlong Huang is with the School of Intelligent Systems Engineering, Sun Yat-sen University
}
}

\begin{document}
 

\newcommand{\sset}{\mathcal{S}}
\newcommand{\frees}{\mathcal{S}_f}
\newcommand{\aset}{\mathcal{A}}
\newcommand{\eps}{\epsilon}
\newcommand{\init}{\rho_0}
\newcommand{\E}{\mathbb{E}}
\newcommand{\R}{\mathbb{R}}
\newcommand{\hS}{\mathbb{S}}
\newcommand{\hP}{\mathbb{P}}
\newcommand{\M}{\mathcal{M}}
\newcommand{\bs}{\vb*{bs}}
\newcommand{\N}{\mathcal{N}}
\newcommand{\G}{\mathcal{G}}
\newcommand{\C}{\mathcal{C}}
\newcommand{\Env}{\mathcal{E}}
\newcommand{\x}{\vb*{x}}

\newcommand{\obj}{o}
\newcommand{\objs}{O}
\newcommand{\pose}{p}
\newcommand{\conds}{C}
\newcommand{\poses}{P}
\newcommand{\initpose}{\pose^0}
\newcommand{\initposes}{\poses^0}
\newcommand{\goalposes}{\poses^g}
\newcommand{\goalpose}{\pose^g}
\newcommand{\bufferpose}{\pose^b}
\newcommand{\cond}{c}
\newcommand{\size}{s}
\newcommand{\cate}{y}
\newcommand{\mask}{m}
\newcommand{\obs}{I_{\objs}}
\newcommand{\func}{f}
\newcommand{\dist}{p_{\func}}
\newcommand{\data}{\textit{D}_{\func}}
\newcommand{\score}{\vb*{\Phi}_{\theta}}
\newcommand{\posespace}{\mathcal{P}}
\newcommand{\condspace}{\mathcal{C}}

\newcommand\mydata[2]{$#1_{\pm#2}$}
\newcommand{\indicator}{\mathbbm{1}}
\newcommand{\loss}{\mathcal{L}}
\newcommand{\trans}{\mathcal{T}}




\def\eg{\emph{e.g}.} \def\Eg{\emph{E.g}.}
\def\ie{\emph{i.e}.} \def\Ie{\emph{I.e}.}
\def\cf{\emph{c.f}.} \def\Cf{\emph{C.f}.}
\def\etc{\emph{etc}.} \def\vs{\emph{vs}.}
\def\wrt{w.r.t. } \def\dof{d.o.f. }
\def\etal{\emph{et al}. }

\maketitle

\begin{abstract}
Visual navigation, a fundamental challenge in mobile robotics, demands versatile policies to handle diverse environments. Classical methods leverage geometric solutions to minimize specific costs, offering adaptability to new scenarios but are prone to system errors due to their multi-modular design and reliance on hand-crafted rules. Learning-based methods, while achieving high planning success rates, face difficulties in generalizing to unseen environments beyond the training data and often require extensive training. To address these limitations, we propose a hybrid approach that combines the strengths of learning-based methods and classical approaches for RGB-only visual navigation. Our method first trains a conditional diffusion model on diverse path-RGB observation pairs. During inference, it integrates the gradients of differentiable scene-specific and task-level costs, guiding the diffusion model to generate valid paths that meet the constraints. This approach alleviates the need for retraining, offering a plug-and-play solution. Extensive experiments in both indoor and outdoor settings, across simulated and real-world scenarios, demonstrate zero-shot transfer capability of our approach, achieving higher success rates and fewer collisions compared to baseline methods. Code will be released at \url{https://github.com/SYSU-RoboticsLab/NaviD}.

\end{abstract}

\section{INTRODUCTION}
Visual navigation~\cite{bonin2008visual, zhang2022survey} is a fundamental challenge in robotics, widely encountered in our daily lives such as unmanned delivery, which requires adaptability to diverse and unseen environments. 
While significant progress has been made in structured environments where pre-built maps are available, path planning in unknown environments based on limited monocular RGB observations without prior information remains challenging. To navigate safely and efficiently, robots must generate collision-free paths in real time, using the available observations to reach their goals despite incomplete scene information.



\begin{figure}[t]
\begin{center}
\includegraphics[width=\linewidth]{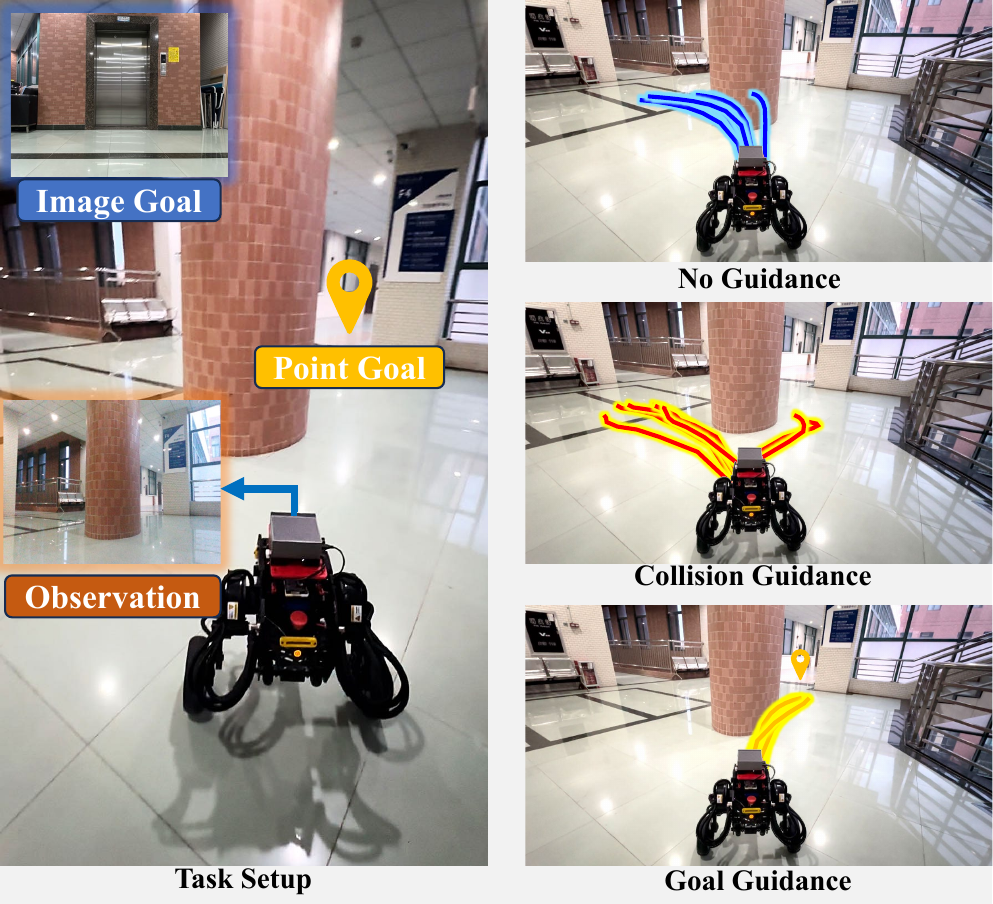}
\end{center}
\vspace{-10pt}
\caption{The robot needs to navigate to destinations (\ie~image goal or point goal) based on given RGB observations. We incorporate collision and goal cost guidance to improve local path generation.}
\label{fig:teaser}
\vspace{-10pt}
\end{figure}


 Previous classical navigation methods typically divide the pipeline into perception, mapping and path planning, with the path planning module often relying on sampling or optimization techniques to minimize designed cost functions~\cite{cao2022autonomous, yang2021real}. While these methods plan effectively and generalize well, they require a modular process to manage real-time perception, maintain a high-quality map, and then search for a valid path. 
 However, the modular design often suffers from information loss at each stage, reducing overall robustness in diverse and complex environments, and leading to impractical modeling of the surrounding environment. 




Recent end-to-end learning approaches~\cite{roth2024viplanner, hoeller2021learning, shah2023vint}, including reinforcement learning and imitation learning, alleviate these issues by exploring integrated pipelines that learn from large-scale data or interactions with simulated environments, directly generating actions or paths from sensory information. This enables the network to develop a prior understanding of various scenes and map them to valid actions. In practice, such learning-based methods can respond quickly and achieve high performance. Despite their success, they face challenges with generalization and stability in out-of-distribution scenes, which are not encountered during training. Additionally, unlike zero-shot classical methods, these approaches require massive high-quality data and substantial training costs. Given the limitations of classical and learning-based approaches, we pose the following question:

\begin{displayquote}
    \textit{How to bridge the gap between classical cost designs and end-to-end learning methods for visual navigation, in a seamless and efficient manner?}
\end{displayquote}


Our key idea is to introduce classical explicit constraints into the inference stage of implicit representations through tailored cost guidance. We first train a conditional generative model (\ie~ diffusion model~\cite{dhariwal2021diffusion}) on large-scale examples of waypoint-based paths with corresponding RGB observations to model the path planning priors. During sampling, the trained diffusion model generates paths by iteratively denoising over $k$ steps. This allows for a unique opportunity to guide the diffusion model by incorporating task-level and scene-specific costs as in the classical planners, ensuring that the paths satisfy scene constraints while retaining their multimodality, as illustrated in Fig.~\ref{fig:teaser}. 




We conduct experiments across various scenarios, including indoor and outdoor scenes, and different goal modalities (\ie~image goal and point goal)~\cite{anderson2018evaluation}, using two robot embodiments to demonstrate the effectiveness of our approach in generating valid paths in real time and deploying them in the real world. Extensive results and analysis showcase that our approach outperforms the baseline in generating collision-free path, particularly in unknown scenes with multiple random obstacles. Ablation studies further indicate that the cost guidance plays an indispensable role in guaranteeing the distribution of generated paths meets scene constraints.


In summary, our key contributions are:

\begin{itemize}
    \item 
    We introduce a novel framework that combines classical and learning-based methods for visual navigation by incorporating the gradients of the designed costs to guide the inference stage of the learned diffusion model.
    \item 
    Our approach generalizes well across diverse scenes from simulation to real-world, and the proposed path selection policy chooses an appropriate candidate from multimodal path distribution to minimize path fluctuations.
    \item 
    We conduct extensive experiments to demonstrate the effectiveness of our approach in generating multimodal collision-free paths and real-world deployment.
\end{itemize}

\section{RELATED WORK}

\subsection{Visual Navigation}
Navigation has been extensively explored in mobile robotics. Classical navigation methods typically frame navigation as a geometric problem, decomposed into two stages: \textit{i)} perceiving and mapping the surroundings using SfM or SLAM~\cite{sfm, ORBSLAM3, Torsten17, chaplot2020learning}, and \textit{ii)}planning a collision-free path to the target based on sampling or optimization~\cite{blosch2010vision, cummins2007probabilistic, paden2016survey}. In the case of using visual input for mapping, methods like \cite{ORBSLAM3, qin2018vins, ren2025layer} extract features from visual observation to perform simultaneous mapping and localization~\cite{cadena2016past}.

More recent works shift research interests towards developing end-to-end policies that directly infer  actions from sensory information (\ie, RGB, depth, etc.)~\cite{kahn2021badgr, loquercio2021learning, ye2021auxiliary, zhu2020vision, chaplot2020neural, ding2024opg}.
These works demonstrate remarkable performance, offering opportunities to learn semantic priors for goal-directed exploration~\cite{gervet2023navigating}, while they require large datasets and extensive training to be transferable to new scenes.

In the case of RGB-only input, \textit{visual} navigation in unseen environments without prior information (\ie, GPS, position, map, etc.) still faces significant challenges. Zhu et al.~\cite{zhu2017target} employ Reinforcement Learning to address target-driven visual navigation, aiming to search for the target in small indoor scenes based on a given image of the target. \cite{mousavian2019visual} and \cite{mayo2021visual} enhance visual representation by incorporating semantic segmentation and spatial attention techniques. ViNT~\cite{shah2023vint} proposes a foundation model with topological graphs for long-horizon visual navigation.
However, these learning-based approaches face challenges with generalizability and reliability in unseen scenarios not covered during training.
In contrast, our method seeks to integrate the strength of classical methods with end-to-end learning, enabling generalizable and reliable performance across various visual navigation tasks.


\subsection{Diffusion for Planning}
Diffusion models have emerged as powerful generative models with stable training characteristics and have demonstrated remarkable multimodal generative capabilities across various robotics domains~\cite{zhang2024generative, zeng2024lvdiffusor, wu2022targf, ren2025navi}. In planning and control, Janner et al.~\cite{janner2022diffuser} leverage diffusion models to directly infer high-dimensional trajectories within a given environment. Diffusion policy~\cite{chi2023diffusionpolicy} further explores the application of diffusion models to learn visuomotor control policies for behavioral cloning.

Recent works also show the potential of diffusion models in the context of visual navigation. In particular, ViNT~\cite{shah2023vint} uses diffusion as a subgoal proposal model to generate subgoal images, while NoMaD~\cite{sridhar2024nomad} employs diffusion, similar to Diffusion Policy~\cite{chaplot2020learning}, to directly infer multimodal actions conditioned on visual observation. Unlike purely end-to-end methods, we explore a hybrid approach that integrates scene-specific explicit geometric and task-level constraints with implicit local path modeling from a pre-trained diffusion model to generate valid paths that meet specific task requirements.

\begin{figure*}[t]
\begin{center}
\includegraphics[width=\linewidth]{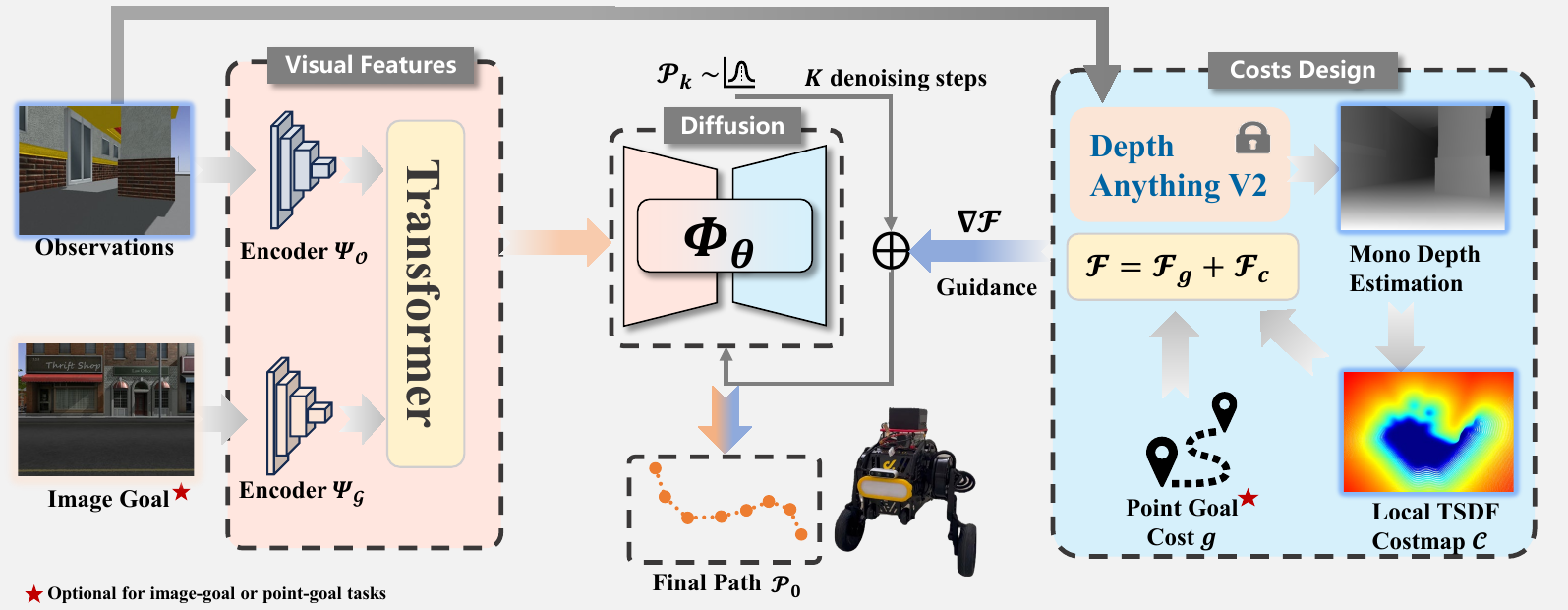}
\end{center}
\vspace{-10pt}
\caption{\textbf{Pipeline overview:} RGB observations and the image goal are processed through two encoders, $\Psi_\mathcal{O}$ and $\Psi_\mathcal{G}$, then fed to transformer, serving as a condition for the diffusion model. The gradient of designed cost function $\nabla \mathcal{F}$ is incorporated at each denoising step to guide the local path generation. For long-horizon navigation, a high-level policy, such as a topological map, is used to provide subgoals, supporting both image and point goals.}
\label{fig:pipeline}
\vspace{-15pt}
\end{figure*}

\section{METHOD}
\textbf{Task Description:} In this work, we aim to design a local path planning policy for visual navigation. The robot is provided with RGB sequences $\mathcal{O} = \{\textbf{I}_t\}^T_{t=T-s}$ 
from past moving observations, the objective is to generate future waypoint-based path $\mathcal{P} = \{W_t\}^{T+n}_{t=T}$ to guide the robot to reach the goal. We consider two types of goals: \textit{i)} RGB image goal $G_I$ and \textit{ii)} Point goal $G_P$. The policy has access to goal information and adaptively navigates to destinations by offering safe, reasonable and collision-free paths.

\textbf{Overview:} We formulate the local path planning as a conditional generative modeling problem Sec.~\ref{sec:guided diffusion} and train a diffusion model $\boldsymbol{\Phi}_\theta$ to learn a prior $p_\theta$ from a large-scale dataset of paired paths and RGB observations $\mathcal{D} = \{(\mathcal{P}_i, \mathcal{O}_i)\}^n_{i=1}$. The proposed pipeline introduced cost guidance Sec.~\ref{sec:cost} into the diffusion process, as illustrated in Fig.~\ref{fig:pipeline}. We aim to sample a group of path candidates via the \textit{cost-guided} diffusion model $\boldsymbol{\Phi}_\theta$, conditioned on the observations $\mathcal{O}$ (and image goal $G_I$ if applicable). Specifically, we construct task-level goals and scene-specific constraints of the path as differentiable costs $\mathcal{F}(\mathcal{P}; \mathcal{O})$, which are used to iteratively guide the reverse denoising process of the diffusion model $\boldsymbol{\Phi}_\theta$ using the gradients $\nabla \mathcal{F}(\mathcal{P}; \mathcal{O})$. Finally, we enhance the performance of the generated path candidates under specific constraints and select the optimal path for implementation based on a path estimator Sec.~\ref{sec:implementation}.

\subsection{Cost-guided Diffusion Model}
\label{sec:guided diffusion}


Given specific well-designed differentiable costs, an intuitive approach would be to perform gradient descent directly on the waypoint-based path. However, the multimodal nature of path distributions can easily cause gradient descent approaches to get stuck in local minima and also make it infeasible to train a regression-based model. Hence, we distill the collected dataset into a conditional generative model $\Phi_\theta$. 
Inspired by classifier guidance~\cite{dhariwal2021diffusion}, we incorporate the designed cost function to guide the diffusion process in the sampling stage. This approach steers the generation towards paths that satisfy specific constraints, enabling a balance between multimodal diversity and scene-specific guidance.

\subsubsection{Diffusion Model}
We employ diffusion models \cite{ho2020denoising, SDEScoreMatching}, a class of probabilistic generative models known for its stable training and promising capabilities in conditional generative modeling, to model the conditional distribution $\dist (\mathcal{P} | O)$ by predicting the noise added to a sample. The training process involves sampling data points from the dataset. For each pair of image observations $\mathcal{O}$ and path $\mathcal{P}$, we initiate a continuous diffusion process $\{\mathcal{P}(t)\}^1_{t=0}$ over the time parameter $t \in [0, 1]$. In particular, we randomly sample a time step $t$ and sample a Gaussian noise $\mathbf{\epsilon}_t$ adding to $\mathcal{P}_0$ to produce perturbed sample $\mathcal{P}_t$. Image observations $\mathcal{O}$ are considered as a condition and the following loss function is defined as:

\begin{equation}
\mathcal{L}=\operatorname{MSE} (\mathbf{\epsilon}_t, \Phi_\theta(\mathcal{O}, \mathcal{P}_t, t)),
\label{mse loss}
\end{equation}
when minimizing the objective $\mathcal{L}$, the optimal noise prediction network $\Phi^*_\theta( \mathcal{P}, t | \mathcal{O})$ approximates the gradient field of noise $\nabla E(\mathcal{P})$.


In the test phase, to generate paths from the trained diffusion model $\Phi_\theta$, we employ DDPM~\cite{ho2020denoising} to perform iterative denoising in $t$ steps, starting from a perturbed sample $\mathcal{P}_t$ which sampled from Gaussian noise, and continuing until the final noise-free sample $\mathcal{P}_0$ is obtained, as detailed in the following equation. 

\begin{equation}
\mathcal{P}_{t-1} = \alpha(\mathcal{P}_t - \gamma \Phi_\theta(\mathcal{O}, \mathcal{P}_t, t) + \mathcal{N}(0, \sigma^2 I)),
\label{step update}
\end{equation}
where $\mathcal{N}(0, \sigma^2 I)$ denotes the Gaussian noise added at each iteration, and $\alpha, \gamma, \sigma$ are noise schedule of the function, can be considered as hyperparameters in gradient descent process.

\subsubsection{Cost-guided Sampling}
\label{sec:sampling}
During the sampling stage, we incorporate scene-specific and task-level cost functions $\mathcal{F}(\mathcal{P}; \mathcal{O})$ to guide reverse diffusion process toward desired waypoint paths that satisfy specific constraints. Inspired by~\cite{dhariwal2021diffusion, xu2024dynamics, saha2024edmp}, we extend classifier-based guidance by utilizing explicit cost representations, termed as cost guidance. Specifically, the classier gradient $\nabla p_\phi$ is replaced with the gradient of cost function $\nabla \mathcal{F}(\mathcal{P}; \mathcal{O})$, which is iteratively computed and added to guide the intermediate paths $\mathcal{P}_t$ predicted by the diffusion model at the $t^{\text{th}}$ time step. The detailed proposed procedure is summarized in Algorithm~\ref{alg:costguided}. 


\begin{algorithm}[b]
\caption{Cost guided reverse diffusion sampling, given a diffusion model $\score$, designed objective $\mathcal{F}(\mathcal{P}; \mathcal{O})$ }
\label{alg:costguided}
\begin{algorithmic}[1]
    \State \textbf{Input:} designed objective $\mathcal{F}(\cdot)$, Covariance schedule $\Sigma_t$, and gradient scale $s_t$
    \State \textbf{Initialization:} Learned diffusion network $\score$, received visual observation $\mathcal{O}$
    \State $\mathcal{P}_T$ $\leftarrow$ sample from $\mathcal{N}(0, \mathbf{I})$
    \For{\textbf{all} $t=$  $T$ \textbf{to} 1}
    \State $\mathcal{P}_{t-1} \sim \mathcal{N}(\score (\mathcal{P}, t | \mathcal{O}) + s_t \nabla_{\mathcal{P}} \mathcal{F}(\mathcal{P}; \mathcal{O}), \Sigma)$
    \EndFor
    \State \textbf{Return} $\mathcal{P}_0$
\end{algorithmic}
\end{algorithm}


\subsection{Cost Guidance}
\label{sec:cost}

Sec.~\ref{sec:sampling} introduces gradients computed from a designed cost function to guide the path sampling process. The cost function, which evaluates the quality of the generated path, consists of two differentiable components: the goal cost $\mathcal{F}_g$ and the collision cost $\mathcal{F}_c$.

\subsubsection{Point-goal Cost Guidance}
For tasks where the robot needs to navigate to a point goal, the generated path distribution should be directed towards specific points. Therefore, we define the goal cost $\mathcal{F}_g$ as the Euclidean distance between the final waypoint of the path $\mathcal{P}$ and the point goal $G_p$, which is formulated as follows:

\begin{equation}
\mathcal{F}_g (\mathcal{P}) = \| W_0 - G_p \|^2 \quad W_0 \sim \mathcal{P},
\label{eq:goal cost}
\end{equation}
where $W_t^0$ denotes the last waypoint in the intermediate path generated by the diffusion model in the sampling stage.

\begin{figure}[t]
\begin{center}
\includegraphics[width=\linewidth]{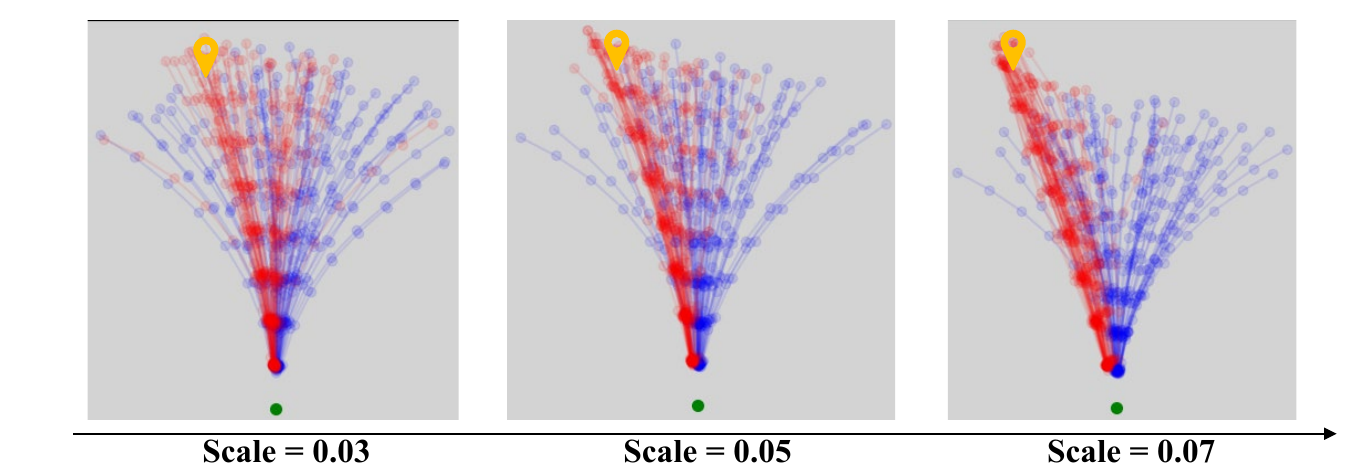}
\end{center}
\vspace{-15pt}
\caption{Effect of different guide scale: The guidance scale increases from left to right, we sample 50 paths with guidance (red) and 50 paths without guidance (blue) for each scale.}
\label{fig:Guide Scale}
\vspace{-12pt}
\end{figure}

The effect of goal guidance in diffusion process is shown in Fig.~\ref{fig:Guide Scale}. From left to right, as we increase the guidance scale $s$, more goal cost guidance is provided, improving performance in reaching the designated point goal but reducing the diversity of the generated paths. Thus, we can balance diversity and precision by adjusting the scale.

\subsubsection{Collision Cost Guidance}
\begin{figure}[t]
\begin{center}
\includegraphics[width=\linewidth]{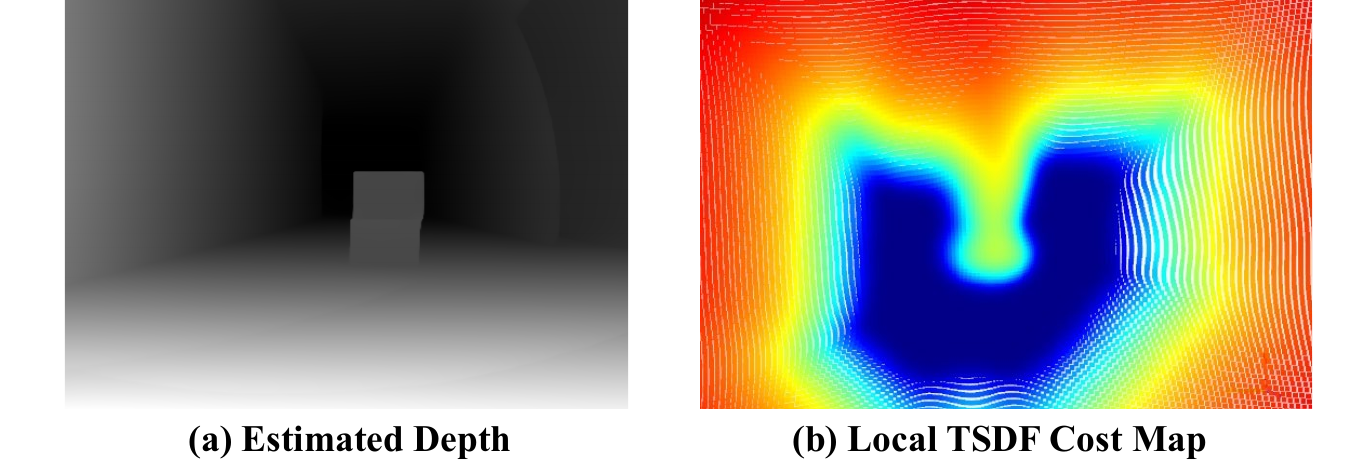}
\end{center}
\vspace{-15pt}
\caption{Example estimated depth and its local TSDF cost map generated from RGB observation in the Stanford 2D-3D-S environment.}
\label{fig:local cost map}
\vspace{-10pt}
\end{figure}

To enhance collision avoidance performance in navigation, a straightforward way involves using depth information from sensors like depth cameras or LiDAR to map the surroundings and check if the path collides with or is too close to obstacles. However, in visual navigation using only RGB input, ground truth depth information is not available. We address this by employing the state-of-the-art monocular depth estimation method, Depth Anything V2~\cite{depth_anything_v2}, to estimate depth from RGB observations. Then, based on the estimated depth image, we reconstruct the surroundings in real-time and build a local Truncated Signed Distance Function (TSDF)~\cite{newcombe2011kinectfusion} to label the distance to the surface of the obstacles for each position in the environment. The local TSDF is then smoothed with a Gaussian filter to make it differentiable, creating a cost map $\mathcal{C}$ with non-negative cost values.
An estimated depth image and the corresponding local TSDF cost map are visualized in Fig.~\ref{fig:local cost map}, showing the effectiveness of this method in reconstructing the nearby surroundings. To adapt to various robot embodiments, the collision cost $\mathcal{F}_c$ accounts for not only each waypoint on the path $\mathcal{P}$, but also points perpendicular to the path at a distance $\mathbf{\sigma}_R$, which represents half the robot's width. All the points are projected on the cost map to obtain their respective cost values. The collision cost $\mathcal{F}_c$ is formulated as follows:


\begin{equation}
\mathcal{F}_c (\mathcal{P}) = \sum_{t=1}^n k_t 
[\mathcal{C}(W_t) + \mathcal{C}(W_t+\mathbf{\sigma}_R) + \mathcal{C}(W_t-\mathbf{\sigma}_R)],
\label{eq:collision cost}
\end{equation}
where $W_t$ denotes all waypoints in the path, $\mathbf{k}$ represents impact factors of costs for each waypoint. 

Overall, the differentiable path cost $\mathcal{F}$, as the guidance for the diffusion, is formulated as a combination of goal cost $\mathcal{F}_g$ and collision cost $\mathcal{F}_c$:

\begin{equation}
\mathcal{F} (\mathcal{P}) = \alpha \mathcal{F}_g (\mathcal{P})+ \beta \mathcal{F}_c (\mathcal{P}),
\label{eq:total cost}
\end{equation}
where $\alpha, \beta$ are hyperparameters to adjust the scale of each term in the cost.


\subsection{Path Selection From Generated Candidates}
\label{sec:implementation}
The nature of the diffusion model allows it to generate multimodal paths given an observation. Abrupt path selection between consecutive time steps can lead to unstable trajectories and planning failures. As shown in Fig.~\ref{fig:teaser}, when encountering obstacles, the path distribution becomes overly deviated, causing the generated path to rapidly switch between the left and right distributions within a short time, which may lead to planning failure. To alleviate this issue, ensuring the consistency and smoothness of the path is essential. 

\textbf{Consistency}: To ensure decision consistency, the robot's driving direction should remain stable across consecutive time steps under similar perception and target conditions, avoiding rapid fluctuations. Let $\mathcal{S}^t = \{\mathcal{P}^t_0, \mathcal{P}^t_1, ..., \mathcal{P}^t_n \}$ represent the $n$ paths generated at time $t$, and $\mathcal{P}^h$ denote the path chosen at a previous time step. Let $\mathcal{V}=\{\mathcal{P}|\delta(\mathcal{P}_t, \mathcal{P}^h)<\epsilon, \mathcal{P} \in \mathcal{S}^t\}$, where $\delta$ measures the direction difference, $\epsilon$ is the difference threshold, the path in $\mathcal{V}$ represents optional actions that are consistent with the path at historical time. The proportion of $\mathcal{V}$ in $\mathcal{S}^t$ ensures that the selected path remains consistent.

\textbf{Continuity}: The path generated by the diffusion model lacks temporal continuity, resulting in hesitations during motion. Consistency-based path selection allows for correcting the current action state using the motion trend of the historical path, providing greater continuity. This is achieved by applying weighted average filtering to the final selected path points from both the historical and current time steps.




\begin{figure*}[t]
\begin{center}
\includegraphics[width=\linewidth]{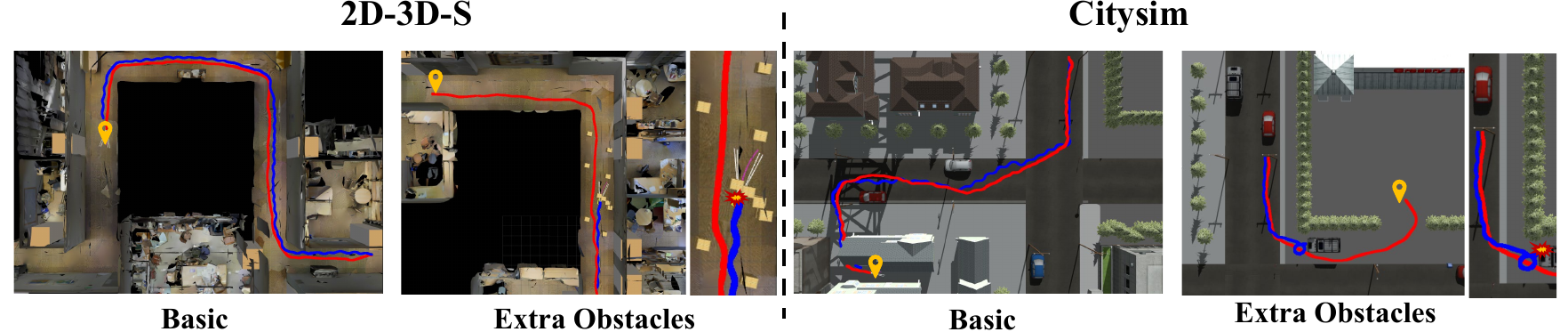}
\end{center}
\vspace{-10pt}
\caption{Qualitative Path Comparison between the proposed NaviDiffusor (Red) and baseline method NoMaD (Blue) in 2D-3D-S and Citysim Environments under Basic and Extra Obstacles Settings. Our method avoids extra obstacles that are not present in the topological map, while the baseline method fails.}
\label{fig:simulation}
\end{figure*}

\begin{table*}[!t]
\centering
\caption{Quantitative comparison between the proposed NaviDiffusor with baselines and ablation}
\setlength{\tabcolsep}{3pt}
    \begin{threeparttable}
        \begin{tabular}{cccccccccccc}
            \toprule
            \textbf{Goal} & \multirow{2}{*}{\textbf{Scene}} & \multirow{2}{*}{\textbf{Method}} & \multicolumn{3}{c}{\textbf{Basic Task}} & \multicolumn{3}{c}{\textbf{Obstacle Task}} & \multicolumn{3}{c}{\textbf{Long-range Task}} \\ 
            \cmidrule(lr){4-6} \cmidrule(lr){7-9} \cmidrule(lr){10-12}
            \textbf{Type} & & & \textbf{Length (m)} & \textbf{Collision} & \textbf{Success} & \textbf{Length (m)} & \textbf{Collision} & \textbf{Success} & \textbf{Length (m)} & \textbf{Collision} & \textbf{Success} \\ 
            \midrule
            \multirow{8}{*}{\rotatebox{90}{\textbf{Image Goal}}} & 
               & ViNT~\cite{shah2023vint} & \textbf{41.1} $\pm$ \textbf{3.172} & 0.66 & 68\%  & 21.4 $\pm$ 0.314 & 0.73 & 42\% & 152.3 $\pm$ 31.590 & 1.02 & 34\% \\ 
              & Indoor & NoMaD~\cite{sridhar2024nomad} & 42.9 $\pm$ 3.283 & 0.37 & 86\%  & 20.3 $\pm$ 0.243 & 0.98 & 58\% & 154.2 $\pm$ 27.381 & 0.74 & 40\% \\ 
             & (2D-3D-S) & Ours w/o guidance & 42.2 $\pm$ 3.281 & 0.05 & 82\%  & 20.0 $\pm$ 0.244 & 0.43 & 52\% & 149.4 $\pm$ 27.257 & 0.83 & 40\% \\ 
             & & Ours & 42.7 $\pm$ 3.278 & \textbf{0.04} & \textbf{100\%}  & \textbf{19.5} $\pm$ \textbf{0.215} & \textbf{0.08} & \textbf{100\%} & \textbf{147.5} $\pm$ \textbf{24.302} & \textbf{0.42} & \textbf{74\%} \\ 
            \cmidrule(lr){2-12}
            & 
               & ViNT~\cite{shah2023vint} & 87.8 $\pm$ 21.597 & 0.22 & 58\% & 67.7 $\pm$ 26.031 & 0.42 & 38\% & 258.1 $\pm$ 64.185 & 0.77 & 20\%\\ 
             & Outdoor & NoMaD~\cite{sridhar2024nomad} & 89.4 $\pm$ 15.348 & 0.13 & 78\% & 68.1 $\pm$ 26.259 & 0.34 & 54\% & 247.6 $\pm$ 67.679 & 0.58 & 36\%\\ 
             & (Citysim) & Ours w/o guidance & 83.5 $\pm$ 13.972 & 0.18 & 78\% & 64.7 $\pm$ 20.846 & 0.38 & 48\% & 230.5 $\pm$ 59.384 & 0.69 & 42\%\\ 
             & & Ours & \textbf{70.8} $\pm$ \textbf{9.561} & \textbf{0.02} & \textbf{98\%} & \textbf{53.3} $\pm$ \textbf{14.279} & \textbf{0.03} & \textbf{92\%} & \textbf{187.6} $\pm$ \textbf{38.186} & \textbf{0.28} & \textbf{68\%} \\ 
            \midrule
            \midrule
            \multirow{8}{*}{\rotatebox{90}{\textbf{Point Goal}}} &
               & ViNT-$P$ & 40.4 $\pm$ 1.399 & 0.12 & 62\% & 21.2 $\pm$ 0.240 & 0.86 & 38\% & 152.3 $\pm$ 29.526 & 0.92 & 50\% \\ 
             & Indoor & NoMaD-$P$ & 41.5 $\pm$ 1.492 & 0.08 & 80\% & 21.7 $\pm$ 0.251 & 0.74 & 46\% & 144.2 $\pm$ 21.121 & 0.55 & 72\% \\ 
             & (2D-3D-S) & Ours w/o guidance & 41.8 $\pm$ 1.486 & 0.07 & 76\% & 21.6 $\pm$ 0.247 & 0.61 & 42\% & 147.5 $\pm$ 21.082 & 0.59 & 74\% \\ 
             & & Ours & \textbf{38.4} $\pm$ \textbf{0.767} & \textbf{0.01} & \textbf{100\%} & \textbf{18.7} $\pm$ \textbf{0.134} & \textbf{0.07} & \textbf{92\%} & \textbf{135.7} $\pm$ \textbf{17.082} & \textbf{0.28} & \textbf{86\%} \\ 
            \cmidrule(lr){2-12}
            & 
               & ViNT-$P$ & 68.1 $\pm$ 5.499 & 0.12 & 72\% & 55.7 $\pm$ 12.589 & 0.68 & 34\% & 216.5 $\pm$ 39.159 & 0.66 & 34\% \\ 
             & Outdoor & NoMaD-$P$ & 69.5 $\pm$ 5.658 & 0.09 & 88\% & 57.9 $\pm$ 13.109 & 0.62 & 42\% & 203.8 $\pm$ 49.561 & 0.57 & 48\% \\ 
             & (Citysim) & Ours w/o guidance & 69.1 $ \pm$ 5.689 & 0.08 & 86\% & 56.8 $\pm$ 12.841 & 0.58 & 38\% & 197.5 $\pm$ 42.982 & 0.55 & 48\% \\ 
             & & Ours & \textbf{64.2} $\pm$ \textbf{1.862} & \textbf{0.01} & \textbf{100\%} & \textbf{48.8} $\pm$ \textbf{3.267} & \textbf{0.02} & \textbf{86\%} & \textbf{169.5} $\pm$ \textbf{25.349} & \textbf{0.21} & \textbf{82\%} \\ 
            \bottomrule
        \end{tabular}
        \label{tab:table1}
        \begin{tablenotes} 
    	\item $-P$ represents the extension of the baseline models to accommodate point-goal inputs.
            \item \textbf{50 trials} of all methods are conducted for each task across all scenes.
        \end{tablenotes} 
    \end{threeparttable} 
    \vspace{-16pt}
\end{table*}

\section{EXPERIMENTS}
In this section, we comprehensively evaluate our method through both simulated and real-world experiments across two goal modalities and three difficulty levels, in both indoor and outdoor environments. In the following sections, an overview of the task setups, evaluation metrics and experiment results are provided.

\subsection{Environmental Setup}

\textbf{Dataset:} For a fair comparison, our method and all baseline methods use the same dataset for training. Following~\cite{sridhar2024nomad}, the training data includes examples collected from various environments and across different robotic platforms, including RECON \cite{shah2021rapid}, SCAND \cite{karnan2022scand}, GoStanford \cite{hirose2019deep}, and SACSoN \cite{hirose2023sacson}. In particular, the dataset comprises image sequences of successive frames, accompanied by the corresponding positional data. 

\textbf{Model Training:} The training process is managed using the AdamW optimizer with a learning rate scheduler, training with a batch size of 256. The training procedure is performed on a single NVIDIA RTX TITAN for around 39 hours to converge. The number of steps $k$ is configured to 10 in the sampling process.

\textbf{Baselines:} We compare our work against two SOTA baselines (NoMaD~\cite{sridhar2024nomad}, ViNT~\cite{shah2023vint}) in image-goal navigation tasks. Following \cite{shah2023vint}, we extend these two baselines to accommodate point goal input by adding linear layers and activation functions, which map the input target coordinates to their shared token space.

\textbf{Metrics:} We report three metrics for evaluation: Length, the mean and variance of the path length for successful tasks; Collision, the average number of collisions per trial. Success, the success rate under identical conditions, the trial is terminated and marked as a failure if the robot fails to reach the destination or becomes stuck due to a collision, exceeding the time limit.



\textbf{Experiment Setup}: We compared our method with baselines and ablation in both indoor and outdoor simulation environments, demonstrating the effectiveness of our approach. Moreover, we applied our method on the robot to showcase its performance in real-world applications. The model operates on Nvidia Jetson Orin AGX deployed on the robot with RGB-only input. 
For image target tasks, we only use the collision cost guidance. For positional target tasks, we employed both goal cost and collision cost together for guidance. Path selection was applied in both tasks. 
We set the default $\alpha$=0.3, $\theta$=$\pi/4$, the weight of the collision cost is 0.006, and the weight of the goal cost is 0.03. For the collision cost, non-uniform weight scaling was applied, giving each point an additional weight coefficient, linearly increasing from 0 to 1 from near to far distances. In the real world experiment, the linear velocity and angular velocity of the robot are 0.5m/s and 0.4rad/s respectively.


\begin{figure*}[t]
\begin{center}
\includegraphics[width=\linewidth]{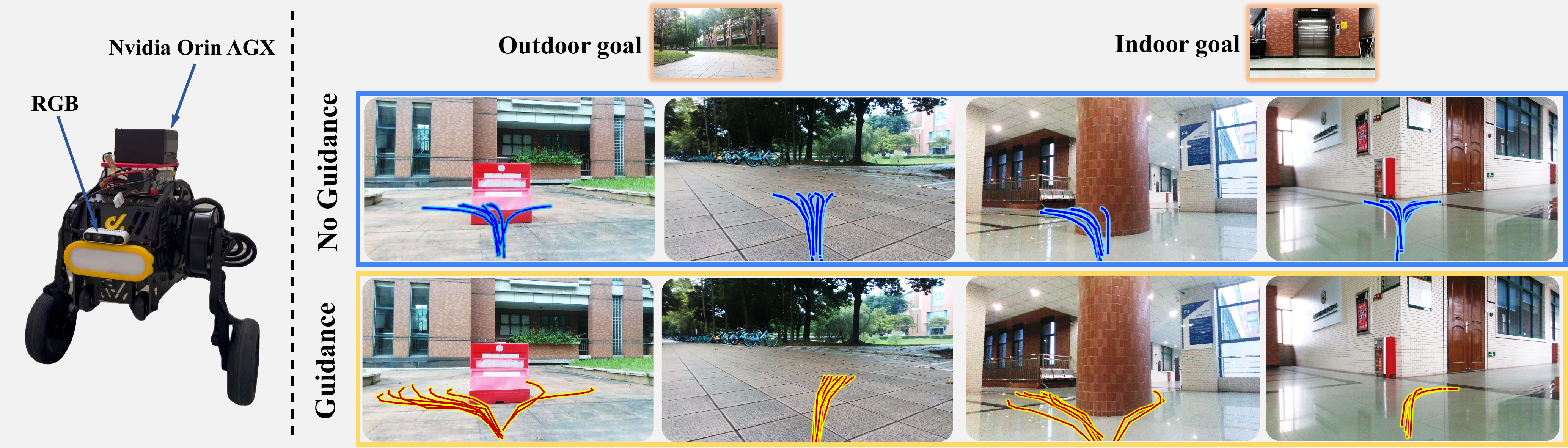}
\end{center}
\vspace{-10pt}
\caption{Qualitative results of real-world experiments with wheeled-leg robot in outdoor and indoor scenarios. Four planning events are visualized, both with guidance (yellow) and without guidance (blue).}
\label{fig:Real world}
\vspace{-15pt}
\end{figure*}

\subsection{Simulation Experiments}
We conduct simulated experiments in two types of scenarios (Fig.~\ref{fig:simulation}): \textit{i)} Indoor (Stanford 2D-3D-S \cite{armeni2017joint}) \textit{ii)} Outdoor (Gazebo citysim \cite{koenig2004design}). In each scenario, we consider both the basic navigation task and the Long-range task with a farther goal, where no additional obstacles are introduced during testing, as well as the more challenging Obstacle task, which includes random new obstacles unknown in the topology and training datasets.
As shown in Fig.~\ref{fig:simulation}, both methods demonstrate comparable performance in basic tasks without obstacles. In the case of challenging tasks with random obstacles, the proposed guided method consistently avoids the collision and achieves the destination, while the baseline is often stuck by random unknown obstacles.

Table.~\ref{tab:table1} presents the results of a further comprehensive evaluation. We evaluate our proposed method with baseline methods and ablation quantitatively by running 50 trials of all alternatives for each task in two types of scenarios.


\subsubsection{Image-goal navigation}

In the Image Goal task for indoor scenarios, NaviDiffusor performs exceptionally well. Without guidance, it achieves an 82\% success rate, close to NoMaD’s 86\%. With guidance, NaviDiffusor reaches 100\% success, surpassing all other methods. In obstacle tasks, it reduces collisions significantly, with an average of just 0.08, while maintaining a 100\% success rate. In long-distance tasks, NaviDiffusor excels in path planning, achieving the best results with the lowest collision and highest success rate.

In outdoor scenarios, NaviDiffusor also shows clear advantages. In the basic task, it achieves a 98\% success rate with guidance, outperforming others with shorter path lengths. In obstacle tasks, it maintains a 100\% success rate with just 0.08 collisions, far better than NoMaD’s 54\%. In long-distance tasks, NaviDiffusor remains stable and efficient, achieving a 68\% success rate, leading other methods in complex environments. The path selection policy allows our method to demonstrate a more substantial improvement in path length compared to other methods, as evidenced by the enhanced path stability illustrated in Fig. ~\ref{fig:simulation}.



\subsubsection{Point-goal navigation}
NaviDiffusor also performs excellently in the Point Goal task. In both the basic and obstacle tasks in indoor scenarios, it achieves a 100\% success rate, with path planning accuracy and robustness significantly better than NoMaD and other baseline methods. In the long-distance task in outdoor scenarios, NaviDiffusor maintains an 82\% success rate, with lower collision rates and superior path planning quality compared to the other methods.

Overall, NaviDiffusor performs excellently across different scenarios and tasks, especially in obstacle-dense and long-distance navigation tasks, where its success rate and collision avoidance capabilities significantly outperform baseline methods like NoMaD. This indicates that NaviDiffusor has strong adaptability and robustness in solving complex navigation tasks, making it an efficient and reliable path planning method. It is noteworthy that in the ablation study, our method without cost guidance shows a significant drop in performance, especially in challenging scenarios.

\subsection{Real-world Experiments}

The following experiments show the effectiveness of our method in real-world scenarios using a wheeled-leg robot Diablo~\cite{liu2024diablo} and Jackal, both equipped with an Intel Realsense D435i only for RGB observations. 

As shown in Fig. \ref{fig:Real world}, guidance significantly improves path planning. Without guidance, the robot's path (blue curve) is more uncertain, with deviations especially pronounced outdoors. In complex indoor environments, paths are inefficient and collision-prone, though still reaching the target. This indicates that NaviDiffusor can plan feasible paths without guidance, but with reduced stability and efficiency.

With guidance, the robot follows more direct, efficient paths in both settings, reducing divergences and collisions. The improvements are especially clear in outdoor scenarios, where the robot moves more intuitively and accurately towards the target. Guidance significantly boosts NaviDiffusor's performance in challenging unknown environments.

\section{CONCLUSION}
In this work, we explore a hybrid mechanism that combines the strengths of classical and learning-based methods. Specifically, the proposed NaviDiffusor learns the priors over large-scale valid paths with paired RGB observations and directly guided by the proposed task-level and scene-specific cost designs at the inference stage. This approach leverages the generalization and robustness of classical methods to facilitate the diffusion model in generating paths satisfying diverse constraints. 
Additionally, this approach can generate multimodal paths for given observations, thereby facilitating optimal path selection with a specific high-level policy, which is crucial for deployment in real robotic systems. 
Our results, including real-world experiments, demonstrate remarkable generalization and reliable capability across more complex new scenes without finetuning. 

While the experiments show the effectiveness of this guidance framework, it still requires sophisticated cost function design. Future work could explore intelligent approaches to cost guidance design and parameters optimization.

{
\bibliographystyle{IEEEtran}
\bibliography{IEEEabrv,reference}
}

\end{document}